\documentclass[sigconf]{acmart}

\usepackage{color}
\usepackage{colortbl}
\usepackage{booktabs}

\usepackage[utf8]{inputenc}
\usepackage{CJK}
\usepackage{multirow}
\newcommand{\kaiti}[1]{\begin{CJK*}{UTF8}{gkai} #1 \end{CJK*}}

\AtBeginDocument{%
  }

\setcopyright{acmlicensed}
\copyrightyear{2024}
\acmYear{2024}
\acmDOI{XXXXXXX.XXXXXXX}

\acmConference[WWW'24]{The Web Conference 2024}{May 13--17,
  2024}{Singapore,Singapore}
  
\acmISBN{978-X-XXXX-XXXX-X/xx/xx}

\begin{document}

\title{General2Specialized LLMs Translation for E-commerce}


\author{Kaidi Chen*, Ben Chen*, Dehong Gao*, Huangyu Dai, Wen Jiang, Wei Ning, Shanqing Yu, Libin Yang†, Xiaoyan Cai†}

\affiliation{%
  \institution{Alibaba Group}
  \streetaddress{699 Wang Shang Road}
  \city{Hangzhou}
  \state{Zhejiang}
  \country{China}
  \postcode{43017-6221}
}

\email{kaidichen@mail.nwpu.edu.cn,chenben.cb@alibaba-inc.com,dehong.gdh@nwpu.edu.cn,}
\email{daihuangyu.dhy@alibaba-inc.com, wen.jiangw@alibaba-inc.com, wei.ningw@alibaba-inc.com,}
\email{yushanqing@zjut.edu.cn,libiny@nwpu.edu.cn,xiaoyanc@nwpu.edu.cn}








\renewcommand{\shortauthors}{Chen et al.}

\begin{abstract}

Existing Neural Machine Translation (NMT) models mainly handle translation in the general domain, while overlooking domains with special writing formulas, such as e-commerce and legal documents. 
Taking e-commerce as an example, the texts usually include amounts of domain-related words and have more grammar problems, which leads to inferior performances of current NMT methods. 
To address these problems, we collect two domain-related resources, including a set of term pairs (aligned Chinese-English bilingual terms) and a parallel corpus annotated for the e-commerce domain. 
Furthermore, we propose a two-step fine-tuning paradigm (named G2ST) with self-contrastive semantic enhancement to transfer one general NMT model to the specialized NMT model for e-commerce. 
The paradigm can be used for the NMT models based on Large language models (LLMs). 
Extensive evaluations on real e-commerce titles demonstrate the superior translation quality and robustness of our G2ST approach, as compared with state-of-the-art NMT models such as LLaMA, Qwen, GPT-3.5, and even GPT-4. 

\end{abstract}





\maketitle
\section{Introduction}\label{sec:intro}
Transformer-based Machine Translation (MT) models have achieved significant process in the general domain, with more training parameters and full richer bilingual parallel corpora~\cite{Vaswani_NIPS2017_Transformer,Liu_TACL2020_mBART,Costa_ArXiv_NLLB,Jinze_arXiv_Qwen}.
Especially for LLMs~\cite{Touvron_ArXiv_LLAMA,Muennighoff_ACL23_BLOOMZ}, peculiar emergence greatly improves their generalization for precise text translation in various sources. 
However, real online tests show these approaches almost all failed for the E-commerce Text Translation (ETT) (e.g. titles, keywords, and detailed descriptions of products). 

The primary cause of this is the difference between the general and e-commerce domains in writing formulas. 
As shown in Tab.~\ref{tab:case}(i), suppliers will attempt to stack the related keywords together as titles of the products. 
This writing formula affects the performances of the NMT approaches based on the general corpus. 

\begin{table*}[t]
\caption{Examples of stacking related words and lacking e-commerce corpora in the fashion domain.}
\newcommand{\tabincell}[2]{\begin{tabular}{@{}#1@{}}#2\end{tabular}}
    \centering
    \footnotesize
    \scalebox{1.08}
    {
    \begin{tabular}{c|c|p{13.5cm}}
    \toprule
    \multicolumn{3}{l}{\bf \uppercase\expandafter{(\romannumeral1)} \bf Stacking of Related Words (product names or attributes)} \\
    \midrule
    \multirow{2}{*}{\bf Case 1} & \bf Chinese & \kaiti{猫产房宠物产房猫狗笼子围栏狗猫窝宠物用品帐篷幼犬幼猫用产房} \\
    \cmidrule(r){2-3}
    &\multirow{1}{*}{\bf English} & Cat Delivery Room Pet Delivery Room Cat and Dog Cage Fence Dog and Cat House Pet Supplies Tent Puppy and Kitten Delivery Room \\
    \midrule
    \multirow{2}{*}{\bf Case 2} & \bf Chinese & \kaiti{气质女装纯色长款针织连衣裙 极简风气质显瘦包臀高领毛衣裙} \\
    \cmidrule(r){2-3}
    &\multirow{1}{*}{\bf English} & Elegant Women's Clothing Long Knitted Dress Minimalist Style Elegant Slimming Sheath Turtleneck Sweater Skirt \\
    \midrule
    \midrule
    \multicolumn{3}{l}{\bf \uppercase\expandafter{(\romannumeral2)} \bf Lack of E-commerce Corpora} \\
    \midrule
    \multirow{3}{*}{\bf Case 1}    &\bf Chinese & \kaiti{\textcolor{red}{厂家批发} 花生米 白沙榨油花生米 特价花生米 \textcolor{red}{代加工}} \\
    \cmidrule(r){2-3}
    &\multirow{1}{*}{\bf English} & \textcolor{red}{Factory Wholesale} Peanut Baisha Oil-pressed Peanut Special Price Peanut \textcolor{red}{Original Equipment Manufacturer}\\
    \midrule
    \multirow{3}{*}{\bf Case 2} & \bf Chinese & \kaiti{日系清新印花短袖衬衫宽松休闲文艺时尚亚麻衬衣\textcolor{red}{一件代发} } \\
    \cmidrule(r){2-3}
    &\multirow{1}{*}{\bf English} & Japanese Style Fresh Printed Short-sleeved Shirt Loose Casual Literary Fashion Linen Shirt \textcolor{red}{One Piece Drop Shipping} \\
    \bottomrule
    \end{tabular}
    }

    \label{tab:case}
\end{table*}

There are two main challenges in ETT. 

1) \textbf{Lack of domain-related resources.} 
As shown in Tab.~\ref{tab:case}(ii), there is a shortage of domain-specific bilingual term pairs, such as "One Piece Drop Shipping", "Processing Customization", and "Original Equipment Manufacturer (OEM)", as well as a scarcity of domain-specific parallel corpora.

2) \textbf{Lack of domain-specialized approaches.} 
Common NMT models are typically trained using large datasets sourced from web-crawled data or daily conversations, which adhere to human grammatical rules. However, in the case of e-commerce text, it is often characterized by keyword stacking and exhibits high language complexity. Consequently, applying conventional NMT methods to e-commerce text may result in keyword omission and duplication. 
These two challenges result in the low accuracy of current NMT approaches in the e-commerce domain. 





\begin{figure}[htp]
\centering
\includegraphics[width=8.4cm]{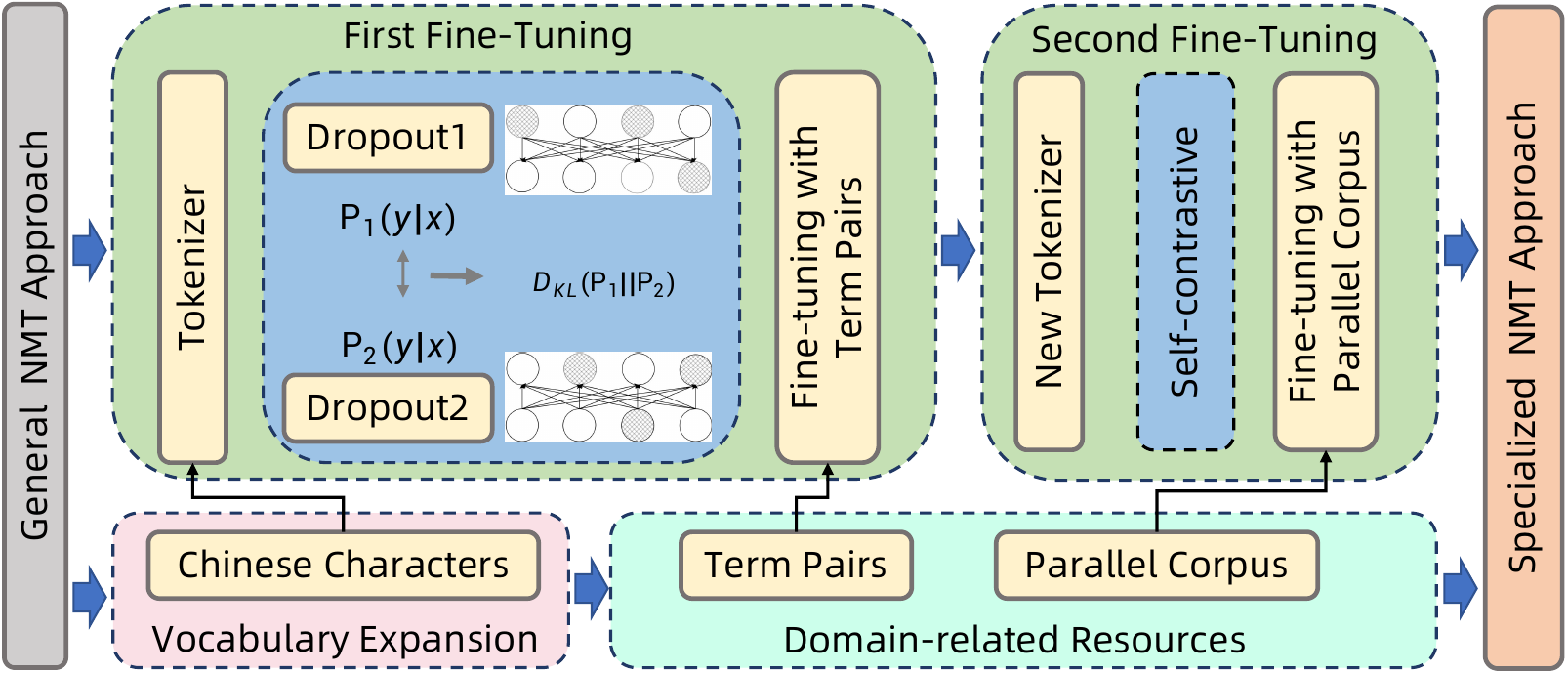} 
\caption{An overview of our G2ST approach. TP = \textit{Terms Pairs}. PC = \textit{Parallel Corpus}. The blue box represents the Self-contrastive module, the gray circle indicates that this node is dropped.}
\label{fig:model}
\end{figure}

In this paper, we propose a General2Specialized Translation approach (G2ST) about Chinese to English, which incorporates \textbf{Two-step Fine-tuning} and \textbf{Contrastive Enhancement}. 
To address the domain-related word issue, we first expand the model vocabulary size (especially, the domain-related word pairs) to avoid the situation of Out-Of-Vocabulary (OOV). 
Meanwhile, we collect Chinese-English bilingual aligned term pairs from Alibaba.com~\footnote{https://www.alibaba.com/} and ChatGPT~\footnote{https://chat.openai.com/}, which are used in the first fine-tuning of the model. 
To address the lack of real e-commerce parallel corpus, we collect and annotate the high-quality Chinese-English parallel corpus of Alibaba.com for second fine-tuning. 
Self-contrastive learning is employed to enhance the robustness and lexical representation capability of the model. The contributions can be summarized as follows: 

\begin{itemize}
    \item We construct two Chinese-English e-commerce translation resources, specifically, aligned bilingual term pairs and parallel corpus tailored to real e-commerce scenes. Both of them are finalized for release, which is believed to benefit the field of ETT greatly.

    \item We propose the two-step fine-tuning paradigm (G2ST) with self-contrastive semantic enhancement to guide one general NMT model to the specialized translation model for e-commerce. 
    
    \item Extensive evaluations on real e-commerce titles demonstrate the superior translation quality and robustness of our approach, as compared with the state-of-the-art NMT models, LLaMA, GPT-3.5, and even GPT-4. 
\end{itemize}

\section{Method}
\label{sec:Method}

\subsection{Domain-related Resources}
Transformer-based models, such as mT5~\cite{Xue_NAACL2021_mT5},  mBART~\cite{Liu_TACL2020_mBART},and NLLB \
~\cite{Costa_ArXiv_NLLB}, often struggle with out-of-vocabulary (OOV) issues in character-centric languages like Chinese and Korean. These models typically use Byte Pair Encoding (BPE)~\cite{Sennrich_ACL2016_BPE} for tokenization, which breaks down unrecognized Chinese characters into the token $\langle unk \rangle$. 
Furthermore, models with restricted vocabularies such as LLaMA and BLOOMZ often split a single Chinese character into multiple tokens, diluting its semantic integrity and hampering translation efficiency. To address these issues, we augment the tokenizer's vocabulary with prevalent e-commerce Chinese characters, mitigating the OOV problem in this domain.

To learn the semantics of newly added characters, we collect the domain-related Chinese terms from Alibaba.com and their English counterparts enhanced by ChatGPT. 
These term pairs include multiple categories such as clothing, maternity, maternal and infant, home furnishings, food, and cosmetics from Alibaba.com. Finally, we created an e-commerce term pairs resource containing 20k pairs of Chinese-English vocabulary. 

In order to adapt to e-commerce scenes, 7000 Chinese titles of products from Alibaba.com are extracted. 
Three annotators with extensive e-commerce experiences are involved in translating these titles into English. 
These e-commerce text pairs constitute domain-specific parallel corpora for e-commerce.

\subsection{General to Special Transfer}
For those Transformer-based models, we develop a two-step paradigm for adapting a general NMT model into the specialized domain model. as shown in Fig.~\ref{fig:model}. 

The first step of fine-tuning is employed to learn the semantics of newly added Chinese characters with the aid of the e-commerce term pairs. 
The second step of fine-tuning is leveraged to learn the writing formulas of the e-commerce experts using the parallel corpus. 
In this way, one NMT approach for general domains is transferred to the specialized domain. 


\subsection{Self-contrastive Semantic Enhancement}

Given the data \(\mathcal{D} = \left \{ (x_{i}, y_{i}) \right \}_{i=1}^{N} \), \((x_{i}, y_{i} )\) is the labeled data pair. For translation, \(x_{i}\) is the source language sentence, and \(y_{i}\) is the target language sentence. The objective of a general model is to minimize the negative log-likelihood loss function, the cross-entropy loss is as follow:

\begin{equation}
    \mathcal{L}_{ce} = -\frac{1}{N}\sum_{i=1}^{n}log\mathcal{P}^{w}(y_{i}|x_{i}) 
\end{equation}

To enhance the output representation, we utilize R-Drop~\cite{Wu_NIPS2021_Rdrop} to reduce the inconsistency existing in training and inference. We feed \(x_{i}\) to go through the forward pass of the network twice and get two distributions of the model predictions, denoted as \(\mathcal{P}_{1}^{w}(y_{i}|x_{i}) \) and \(\mathcal{P}_{2}^{w}(y_{i}|x_{i})\). Due to the dropout operator randomly deactivating units within a model, the two forward passes are essentially conducted using two distinct sub-models, albeit within the same overarching model. In each training step, the R-Drop method aims to regularize the model's predictions by minimizing the bidirectional Kullback-Leibler(KL) divergence between the two output distributions for the same sample, the total KL loss is as follow: 
 
\begin{equation}
\begin{split}
    \mathcal{L}_{KL} = \frac{1}{2N}\sum_{i=1}^{N}(\mathcal{D}_{KL}(\mathcal{P}_{1}^{w}(y_{i}|x_{i})||\mathcal{P}_{2}^{w}(y_{i}|x_{i})) \\
    + \mathcal{D}_{KL}(\mathcal{P}_{2}^{w}(y_{i}| x_{i})||\mathcal{P}_{1}^{w}(y_{i}|x_{i})))
\end{split}
\end{equation}

With the two forward passes, the cross-entropy loss of the model changes to:

\begin{equation}
    \mathcal{L}_{CE} = -\frac{1}{2N} \sum_{i=1}^{N}(log\mathcal{P}_{1}^{w}(y_{i}|x_{i})+log\mathcal{P}_{2}^{w}(y_{i}|x_{i} ))
\end{equation}

Finally, G2ST can be optimized by jointly minimizing the traditional CE loss and contrastive loss:
\begin{equation}
    \mathcal{L} = \mathcal{L}_{CE} + \alpha \mathcal{L}_{KL}
\end{equation}

\noindent where \(\alpha \) is the coefficient weight to control \(\mathcal{L}_{KL}\).

\section{Experiments}
\label{sec:Experiments}

\subsection{Experimental Settings}

\textbf{Dataset and Metrics.} For the e-commerce term pairs, due to their limited length, we include them entirely in the training set and do not perform testing on them. For the e-commerce parallel corpus, we use 5000 of them as the train set and the remaining 2000 as the test set. We evaluate our model on the parallel corpus dataset test set. The metrics we used to measure the translation quality are SacreBLEU~\cite{Post_ACL2018_SacreBLEU}, Rouge-1, Rouge-2, and Rouge-L~\cite{Lin_2004_Rouge}.

\noindent \textbf{Implementation Details.} In the experimental section, we do not make any modifications to the model structure used, keeping it consistent with the default settings. The batch size, initial learning rate, and dropout rate are set to 64, 5e-5, and 0.1, respectively. We utilize Adam~\cite{Kingma_ArXiv2014_Adam} optimizer to optimize our models. For \( \mathcal{L}_{KL}\), we conduct lots of experiments of the coefficient weight \(\alpha \in \left \{ 0.05, 0.10, 0.15, 0.20, 0.30, 0.40, 0.50\right \} \) on NLLB-Distilled and mBART-50, and find that when the coefficient weight \(\alpha\) is set to 0.05, the metrics of models perform best. For Qwen, the original vocabulary includes 7,000 commonly used Chinese characters, which suffices for Chinese-English e-commerce translation needs without necessitating any expansion of vocabulary. For NLLB, the original vocabulary contained Chinese characters of 2575, which cannot meet the needs of the e-commerce field. For example, there are no common characters such as chicken and duck, so we have expanded its Chinese vocabulary to 7000. For mBART-50, we adopt the same approach as NLLB to expand the Chinese vocabulary to 7000.

\begin{table}[t]
\caption{Comparison of STMs and LLMs on the parallel corpus resource. The G2ST(OURS) is based on Qwen-14B. The best is in bold. }
\center
\small
\scalebox{0.88}{
\begin{tabular}{l||r|cccc}
\toprule
\bf Model & \bf Params & \bf SacreBLEU & \bf Rouge-1 & \bf Rouge-2 & \bf Rouge-L \\ \midrule
\multicolumn{5}{l}{~~\textit{Open-source LLMs}}\\
 LLaMA    &  7B   &   25.76  & 62.76 & 40.09 & 62.09  \\ 
 LLaMA&13B    &       27.14  & 64.77 & 42.25 & 64.03  \\ 
 LLaMA2&7B    &           34.98  & 71.16 & 50.71 & 70.47  \\ 
 LLaMA2&13B   &          35.28  & 71.81 & 51.39 & 71.08  \\ 
 Qwen& 1.8B   &  37.18 & 74.19 & 57.78 & 73.54 \\
 Qwen& 7B   &           37.20&	74.45&	57.85&	73.68  \\
 Qwen& 14B   &         38.22&	75.65&	59.51&	74.94 \\
 \midrule
 \multicolumn{5}{l}{~~\textit{Closed-source LLMs}}  \\
 GPT-3.5 & -  &            3.39      & 68.37 & 47.40 & 66.66  \\
 GPT-4   & -  &    6.91 & 73.34 & 53.05 & 71.99   \\ \midrule
 \multicolumn{5}{l}{~~\textit{Specialized Translation Models}}\\
 mT5$_{\textit{base}}$    & 580M     &    13.51 & 46.64 & 23.93 & 46.00    \\ 
 mT5$_{\textit{large}}$  & 1.2B      &    21.12 & 57.61 & 35.50 & 56.90    \\ 
 NLLB-Distilled & 1.3B                     &    35.37 & 69.54 & 49.69 & 68.96    \\ 
 mBART-50$_{\textit{large}}$ & 611M &    36.87 & 71.22 & 50.58 & 70.53     \\ \midrule
 \bf G2ST(OURS)    & 14B &  \bf 39.85 & \bf 76.89 & \bf 60.86 &  \bf 75.74   \\ 
\bottomrule
\end{tabular}
}
\label{tab:results-Comparison experiment}
\end{table}

\subsection{Main Results}

As shown in Tab.~\ref{tab:results-Comparison experiment}, we compare our G2ST with open-source LLMs, closed-source LLMs, and Specialized Translation Models (STMs). 

For both open-source and closed-source LLMs, we employ the prompt ``Translate this sentence from Chinese to English'' to instruct the LLMs to function as translation models, thereby facilitating the execution of sentence translation tasks.
Results show that LLaMA2 outperforms LLaMA by approximately 10 points on various metrics and increasing the parameters from 7 billion to 13 billion only leads to a marginal improvement of 1-2 points in the performance metrics. 
The performance of Qwen-1.8B is comparable to Qwen-7B, and Qwen-14B is the best-performing model among all baselines. 
ChatGPT excels in Rouge scores but has lower SacreBLEU scores at 3.39/6.91. Analysis shows it translates product keywords accurately but adds irrelevant words, reducing its SacreBLEU scores.
For STMs, NLLB-Distilled and mBART-50$_{\textit{large}}$ are comparable to LLaMA2, with impressive results across various metrics.
We find that the current translation capabilities of LLMs are comparable to or even exceed STMs.
We conduct G2ST based on Qwen-14B, NLLB-Distilled, and mBART-50$_{\textit{large}}$, and ultimately obtain the best model based on Qwen-14B, which gets +1.63 SarceBLEU, +1.24 Rouge-1, +1.35 Rouge-2, and +0.80 Rouge-L compared to other SOTA models. 

\begin{table*}[t]
\caption{Ablation study on domain-related resources. EV = \textit{Expand Vocabulary}, TP = \textit{Term Pairs}, PC = \textit{Parallel Corpus}. The best is in bold. }
\center
\small
\scalebox{1}{
\begin{tabular}{c||c|c|cc|cc|cccc}
\toprule
\multirow{2}{*}{\bf Model} &\multirow{2}{*}{\bf No.} &\multirow{2}{*}{\bf EV} & \multicolumn{2}{c|}{\bf TP} & \multicolumn{2}{c|}{\bf PC} & \multirow{2}{*}{\bf SacreBLEU} & \multirow{2}{*}{\bf Rouge-1} & \multirow{2}{*}{\bf Rouge-2} & \multirow{2}{*}{\bf Rouge-L} \\
         &  &     & \(\mathcal{L}_{CE}\) & \(\mathcal{L}_{KL}\) & \(\mathcal{L}_{CE}\) & \(\mathcal{L}_{KL}\) &&&&\\  \midrule
\multirow{4}{*}{\bf NLLB-Distilled} & A  &   &     &       &         &         &  1.97  & 48.42 & 24.31 & 47.08  \\
    & B &   &     &         &       \checkmark       &               &   35.37 & 69.54 & 49.69 & 68.96     \\
    & C &  \checkmark   & \checkmark  & &  \checkmark   &         &   36.83      & 72.67 & 53.27 & 72.05  \\ 
    & D &  \checkmark   &\checkmark& \checkmark & \checkmark & \checkmark& \bf 38.53 & \bf 73.75 & \bf 55.02 & \bf 73.13   \\ \midrule
\multirow{4}{*}{\bf mBART-50$_{\textit{large}}$} & A  &   && &    &   & 7.44 & 52.34 & 28.14 & 51.63  \\
    & B &   &     &         &       \checkmark       &             &   36.87 & 71.22 & 50.58 & 70.53     \\
    & C &  \checkmark   & \checkmark  & &  \checkmark   &         & 37.91 & 72.36 & 52.57 & 71.82   \\ 
    & D &  \checkmark   &\checkmark& \checkmark & \checkmark & \checkmark& \bf 37.99 & \bf 73.28 & \bf 53.65 & \bf 72.64  \\
\bottomrule
\end{tabular}
}
\label{tab:results-Ablation experiment}
\end{table*}

\begin{table}[t]
\caption{Ablation study of self-contrastive semantic enhancement on domain-related resources. SSE = \textit{Self-contrastive Semantic Enhancement}. The best is in bold. }
\center
\small
\scalebox{0.95}{
\begin{tabular}{c||c|cccc}
\toprule
\bf Model &\bf SSE & \bf SacreBLEU & \bf Rouge-1 &\bf Rouge-2 & \bf Rouge-L \\  \midrule
\multirow{2}{*}{\bf LLaMA2-7B}   &  &         34.98  & 71.16 & 50.71 & 70.47     \\
&\checkmark &\bf 36.36&	\bf 73.01&	\bf 53.73&	\bf 72.22\\ \midrule
\multirow{2}{*}{\bf Qwen-7B} && 37.20&	74.45&	57.85&	73.68 \\
& \checkmark & \bf 38.43& \bf 76.03& \bf 60.46& \bf 75.27 \\
\bottomrule
\end{tabular}
}
\label{tab:results-Ablation experiment2}
\end{table}

\subsection{Ablation Study}

As shown in Tab.~\ref{tab:results-Ablation experiment}, we conduct extensive analyses on NLLB-Distilled and mBART-50$_{\textit{large}}$ to demonstrate the effectiveness of each component in G2ST. 
For both models, Model B with the strategic vocabulary expansion and focused fine-tuning on term pairs resulted in a tangible enhancement in translation accuracy, as evidenced by the positive gains in both SacreBLEU and the Rouge metrics over Model C. The incorporation of self-contrastive enhancement further propelled the performance, solidifying its role in the model's ability to discern finer nuances in translation tasks.
As shown in Tab.~\ref{tab:results-Ablation experiment2}, we present an ablation study to assess the impact of the Self-contrastive Semantic Enhancement (SSE) on LLaMA2 and Qwen with domain-related resources. It also demonstrates the effectiveness of our self-contrast enhancement method on LLMs. 

\section{Related Work}
\label{sec:format}

Existing LLMs, such as ChatGPT~\cite{chatgpt}, OPT~\cite{Zhang_ArXiv2022_OPT}, BLOOM~\cite{Scao_ArVix2022_Bloom}, LLaMA~\cite{Touvron_ArXiv_LLAMA}, Qwen~\cite{Jinze_arXiv_Qwen} have shown excellent performances in MT~\cite{Jiao_arXiv2023_Chatgpt,Bawden_arXiv2023_Investigating,Zhu_ArXiv_Multilingual}. 
~\cite{Jiao_arXiv2023_Chatgpt} demonstrates that ChatGPT achieves competitive performance with commercial translation products such as Google Translate for high-resource European languages and has become a good translator with GPT-4 engine. 
~\cite{Bawden_arXiv2023_Investigating} and ~\cite{Zhu_ArXiv_Multilingual} show that LLMs like ChatGPT and BLOOM possess multilingual translation capabilities, surpassing even state-of-the-art online translation engines. 
In addition to LLMs, there are also models specifically designed for translation. Since the advent of Transformer~\cite{Vaswani_NIPS2017_Transformer}, specialized translation models (STMs) that employ Transformer as the backbone, such as mT5~\cite{Xue_NAACL2021_mT5}, mBART~\cite{Liu_TACL2020_mBART}, M2M~\cite{Fan_JMLR2021_M2M}, and NLLB~\cite{Costa_ArXiv_NLLB}, are constantly emerging. All of them are pre-trained on large amounts of data and suitable for multilingual translation. 
However, these studies are conducted in a general field without considering the differences in specialized fields.

\section{Conclusion}
\label{sec:Conclusion}
In this paper, we propose the G2ST approach to convert a general NMT model to the specialized translation model for e-commerce. It is worth mentioning that G2ST can be applied to different models which support the task of MT. 
What's more, we construct two e-commerce Chinese-English translation resources for fine-tuning. 
In the future, we will explore how to extend this approach to multilingual MT. 


\bibliographystyle{ACM-Reference-Format}
\bibliography{acmart}


\end{document}